\PassOptionsToPackage{unicode}{hyperref}
\PassOptionsToPackage{hyphens}{url}
\documentclass[11pt,letterpaper]{article}
\usepackage[margin=0.85in]{geometry}
\usepackage{amsmath,amssymb}
\usepackage{iftex}
\ifPDFTeX
  \usepackage[T1]{fontenc}
  \usepackage[utf8]{inputenc}
  \usepackage{textcomp} 
\else 
  \usepackage{unicode-math} 
  \defaultfontfeatures{Scale=MatchLowercase}
  \defaultfontfeatures[\rmfamily]{Ligatures=TeX,Scale=1}
\fi
\usepackage{lmodern}
\ifPDFTeX\else
\fi
\IfFileExists{upquote.sty}{\usepackage{upquote}}{}
\IfFileExists{microtype.sty}{
  \usepackage[]{microtype}
  \UseMicrotypeSet[protrusion]{basicmath} 
}{}
\makeatletter
\@ifundefined{KOMAClassName}{
  \IfFileExists{parskip.sty}{%
    \usepackage{parskip}
  }{
    \setlength{\parindent}{0pt}
    \setlength{\parskip}{6pt plus 2pt minus 1pt}}
}{
  \KOMAoptions{parskip=half}}
\makeatother
\usepackage{xcolor}
\usepackage{longtable,booktabs,array}
\usepackage{calc} 
\usepackage{etoolbox}
\makeatletter
\patchcmd\longtable{\par}{\if@noskipsec\mbox{}\fi\par}{}{}
\makeatother
\IfFileExists{footnotehyper.sty}{\usepackage{footnotehyper}}{\usepackage{footnote}}
\makesavenoteenv{longtable}
\usepackage{graphicx}
\makeatletter
\def\maxwidth{\ifdim\Gin@nat@width>\linewidth\linewidth\else\Gin@nat@width\fi}
\def\maxheight{\ifdim\Gin@nat@height>\textheight\textheight\else\Gin@nat@height\fi}
\makeatother
\setkeys{Gin}{width=\maxwidth,height=\maxheight,keepaspectratio}
\makeatletter
\def\fps@figure{htbp}
\makeatother
\ifLuaTeX
  \usepackage{selnolig}  
\fi
\usepackage{bookmark}
\IfFileExists{xurl.sty}{\usepackage{xurl}}{} 
\hypersetup{
  hidelinks,
  pdfcreator={LaTeX via pandoc}}

\title{Residual Cross-Model Narrative Convergence under Shared Prompts: Auditing Prompt Controls and Narrative Measurement}
\author{Mehdy Sedaghat Payam\\\small Corresponding author: \href{mailto:mehdyspayam@gmail.com}{mehdyspayam@gmail.com}}
\date{}

\AtBeginEnvironment{longtable}{\small\setlength{\tabcolsep}{4pt}}
\hypersetup{pdftitle={Residual Cross-Model Narrative Convergence under Shared Prompts},pdfauthor={Mehdy Sedaghat Payam}}
\begin{document}
\maketitle
\begin{abstract}

Stories created by LLMs based on the same prompts can be narratively
similar to one another. This similarity may reflect constraints imposed
by the prompts as well as the models\textquotesingle{} writing
tendencies. The current research analyzes 51,240 stories created by five
models in the StoryScope corpus based on 10,248 common prompts. It uses
213 narrative features in seven dimensions to compare cross-model
distances between stories generated from the same prompts with distances
between stories generated from different but semantically matched
prompts. To assess the relationship between narrative features and
prompt content, a semantic relevance mask was evaluated against 420
author-annotated prompt--feature judgments (F1 = .813; near-threshold
accuracy = .621). The equally weighted mean Residual Convergence Index
(RCI) was .409 without masking and .406 with masking. These findings
indicate measurable cross-model narrative convergence relative to a
semantically matched-prompt baseline, with little change following
semantic relevance masking.

\end{abstract}


\section{1. Evaluating similarity under shared prompt
conditioning}\label{evaluating-similarity-under-shared-prompt-conditioning}

A challenge in evaluating texts produced by language models in response
to a prompt stems from the nature of the task itself. If multiple models
produce very similar texts, it may be difficult to determine whether
this similarity is due to the models\textquotesingle{} writing
tendencies or the identical prompts given to them, or both. This is
especially true when assessing creative writing, such as story
generation. Prompts usually describe topics, settings, characters,
conflicts, points of view, or the general direction of the story, and
these instructions may influence the narrative characteristics of the
resulting texts. For this reason, similarities among texts generated in
response to identical prompts cannot, on their own, demonstrate a
tendency towards narrative convergence independent of prompt effects.

This distinction is important for research on the diversity of
AI-generated texts. Doshi and Hauser (2024) found that using AI
assistance can improve evaluations of individual stories while reducing
the diversity of stories produced as a group. In a related study,
Anderson et al. (2024) found that people using LLMs to generate ideas
produced ideas that were more similar in meaning than those produced
using a non-AI creativity-support tool, such as the cards from ``Oblique
Strategies.'' Several studies of AI-generated fiction have also
identified recurring plots, similar emotional trajectories, and familiar
narrative solutions (Tian et al., 2024; Xu et al., 2025; Rettberg \&
Wigers, 2025; Ma et al., 2026). These studies raise questions about
output similarity without isolating how much of it can be attributed to
shared prompts.

However, AI assistance does not necessarily reduce diversity under all
conditions. The extent to which AI reduces diversity may depend on the
particular algorithm and its use in the writing process (Padmakumar \&
He, 2024). For example, Jiang et al. (2025) make a distinction between
repetitive responses to open-ended queries made by one model and
similarities between responses to the same queries by different models.
Likewise, Schaeffer et al. (2026, preprint) question whether the
cross-model homogeneity observed in previous studies depends on the
prompts and the baselines selected for comparison. These debates make
the choice of a comparison baseline especially important.

The dataset used here is the StoryScope corpus, in which five large
language models created stories based on a set of shared prompts. First,
Icompute the distance between stories written by different models using
the same prompt and then compare it with distances between stories
written by different models using distinct but semantically similar
prompts. As a complementary analysis, narrative features assigned the
highest prompt-relevance scores by the semantic mask are excluded. This
methodology does not eliminate prompt effects; rather, it examines how
the comparison changes. Since prompts may include, imply, or not specify
certain features within the story, it is necessary to verify the
semantic mask on the basis of human judgment. The classifications of the
mask are evaluated using 420 pairs of prompts and features from 396
different prompts. Throughout the study, ``residual'' denotes a contrast
relative to the selected matched-prompt controls, not a
prompt-independent tendency of the models.

Specifically, the present study contributes a framework for assessing
how the estimated narrative-similarity contrast depends on prompt
controls and feature measurement, through the following steps: (1) An
audit of 420 prompt--feature pairs across 396 unique prompts by the
author (Annotator A), to examine the constraints imposed by prompts on
narrative features; (2) Investigation of classification errors in the
semantic mask procedure, particularly in borderline cases; (3)
Introduction of the residual convergence index (RCI), based on the
difference between mean cross-model narrative distances under
semantically similar but distinct prompts and under identical prompts,
divided by the former; (4) Application of this method to 51,240 stories,
along with several sensitivity analyses involving the prompt-matching
method, semantic-similarity caliper, dimension weighting,
leave-one-dimension-out tests, and an exploratory ten-dimension
analysis; and (5) Author-reviewed recoding of 400 story records
(Annotator A), followed by direct comparison of the results with
StoryScope\textquotesingle s original feature values. This same-sample
comparison is especially important because it isolates sensitivity to
the extracted feature values.

\section{2. Related work: conditional generation, annotation, and
narrative
resources}\label{related-work-conditional-generation-annotation-and-narrative-resources}

Wenger and Kenett (2026) found that the answers from various language
models to creativity tasks presented in a standardized way were more
similar to each other than human answers were to one another.
Nevertheless, the similarity was measured in responses to creativity
tasks rather than in story features like perspective, chronology, and
characterization. Moreover, automated diversity metrics might miss
distinctions that human readers will notice. Tevet and Berant (2021)
showed that automated diversity metrics do not always coincide with
human judgments of content diversity and that changes produced by
adjusting decoding parameters may affect form without substantially
changing meaning.

The StoryScope method encodes human-written and AI-written stories using
304 narrative features categorized into ten dimensions (Russell et al.,
2026). Other computational approaches have examined individual aspects
of narrative structure, including turning points, plot points, changes
in semantic trajectories, and the functions of narrative parts (Tian et
al., 2024; Xu et al., 2025; Rathje, 2026; Ma et al., 2026). The
advantage of using explicit narrative features is that they allow us to
analyze the semantic connection between the prompt and each feature,
rather than assessing similarity only at the level of entire stories.

Narrative theory makes a distinction between the story, including its
events and characters, and the way the story is presented (Chatman,
1978). In addition, Genette (1980) makes a distinction between narrative
voice and focalization in relation to the questions of who tells the
story and whose point of view determines the information presented in
the narrative. While the StoryScope framework categorizes certain
attributes as falling under the category of Perspective, it should be
kept in mind that these attributes do not constitute complete measures
of theoretical notions of voice and focalization. The same applies to
other categories, including Temporal Structure, Agents, Events, and
Plot. These features can serve as measurable indicators of certain
aspects of narration, rather than complete representations of the
theoretical concepts to which they refer (Moretti, 2013; Piper, 2018).

In order to validate the measurement, it is important to distinguish the
theoretical concept from the indicator used to measure it (Jacobs \&
Wallach, 2021) and to evaluate the applicability of automated
text-analysis techniques to the particular task (Grimmer \& Stewart,
2013). The current study therefore examines two issues: first, whether
the semantic mask identifies narrative features relevant to each prompt;
and second, how StoryScope\textquotesingle s feature values compare with
the author\textquotesingle s recoding of the corresponding story
properties.

\section{3. Data and evaluation
framework}\label{data-and-evaluation-framework}

\subsection{3.1. Corpus and analytic
sample}\label{corpus-and-analytic-sample}

The current study relies on the publicly available StoryScope dataset
(Russell et al., 2026), which comprises 10,272 prompts, AI-generated
stories, information about the original human-written stories, a
taxonomy of narrative types, and narrative features identified in the
stories. The AI-generated stories were produced by the following five
models: GPT-5.4, Claude Sonnet 4.6, DeepSeek V3.2, Kimi K2.5, and Gemini
3 Flash. No additional stories were generated for this investigation;
therefore, these models are treated as five fixed sources of generated
stories.

In total, the narrative feature table from StoryScope contains 10,272
rows each for GPT, DeepSeek, and Kimi, compared with 10,259 rows for
Claude and 10,261 rows for Gemini. Since it was necessary to conduct the
comparison using the same prompts for the five models, I included only
those prompts for which usable narrative feature rows were available for
all five models. In total, 24 prompts (0.234\% of the 10,272 original
prompts) were excluded from the analysis due to missing feature rows for
one or more models, resulting in a total of 10,248 prompts and 51,240
generated stories. The human-written stories were not included in the
comparison since the StoryScope prompts were derived from those stories
rather than independently assigned to human authors.\\
\strut \\
For each prompt used in the analysis, there is one story for each of the
five models. Thus, ten different pairs of models can be compared to each
other within the scope of each particular prompt. The distance between
the stories produced by two models in response to one prompt is
identical irrespective of the order of their enumeration. For this
reason, each of the ten model pairs is counted only once in the
same-prompt comparison. This order invariance does not apply to the
matched-control test. In each particular pair of models, the first one
provides a story produced on the basis of the focal prompt, whereas the
second one provides stories produced in response to semantically similar
but different prompts (see Section 3.4). If the roles of the two models
are reversed, the result is a different matched comparison.

\subsection{3.2. Narrative feature resource and evaluation
boundary}\label{narrative-feature-resource-and-evaluation-boundary}

The narrative features used to compute the level of convergence are
selected through a fixed procedure. Using the taxonomy.json file from
the StoryScope corpus, I include all features belonging to seven
dimensions: Agents (54 features), Social Networks (39), Events (26),
Plot (28), Revelation (25), Temporal Structure (24), and Perspective
(17). This gives a total of 213 narrative features. I exclude three
other dimensions comprising 91 features: Style (39 features), Setting
(27), and Situatedness (25). The list of 213 features and the
prompt--feature combinations remain fixed throughout the procedure. The
semantic mask determines which prompt--feature comparisons are excluded
from the masked calculations on the basis of their relevance to the
prompts. The complete list of selected features, along with the exact
list of 10,248 prompts used for the final computation, is provided in
Online Resource 1. Each feature is identified by its ID, category, name,
identification question, answer type, possible answers, and guidelines
for detecting the feature in a story.

\begin{longtable}[]{@{}
  >{\raggedright\arraybackslash}p{(\columnwidth - 4\tabcolsep) * \real{0.23}}
  >{\raggedright\arraybackslash}p{(\columnwidth - 4\tabcolsep) * \real{0.13}}
  >{\raggedright\arraybackslash}p{(\columnwidth - 4\tabcolsep) * \real{0.64}}@{}}
\caption{Primary narrative feature boundary}\label{tab:1}\\
\toprule\noalign{}
\begin{minipage}[b]{\linewidth}\raggedright
\textbf{Dimension}
\end{minipage} & \begin{minipage}[b]{\linewidth}\raggedright
\textbf{Features}
\end{minipage} & \begin{minipage}[b]{\linewidth}\raggedright
\textbf{Interpretive scope}
\end{minipage} \\
\midrule\noalign{}
\endfirsthead
\toprule\noalign{}
\begin{minipage}[b]{\linewidth}\raggedright
\textbf{Dimension}
\end{minipage} & \begin{minipage}[b]{\linewidth}\raggedright
\textbf{Features}
\end{minipage} & \begin{minipage}[b]{\linewidth}\raggedright
\textbf{Interpretive scope}
\end{minipage} \\
\midrule\noalign{}
\endhead
\bottomrule\noalign{}
\endlastfoot
Perspective & 17 & Narrator position, focalization, access, and
viewpoint organization \\
Social networks & 39 & Relationship structure, social embedding,
interaction, and network organization \\
Agents & 54 & Character roles, agency, goals, interiority, and emotional
organization \\
Events & 26 & Event progression, escalation, causality, and action
sequencing \\
Plot & 28 & Conflict architecture, reversals, resolution, and structural
closure \\
Revelation & 25 & Withholding, suspense, disclosure, mystery, and
information release \\
Temporal structure & 24 & Chronology, ordering, duration, and temporal
discontinuity \\
\end{longtable}

Note: 213 features total. Style, Setting, and Situatedness are excluded
from the primary analysis. The interpretive scopes summarize the aspects
of narration associated with each dimension rather than providing
complete measures of the corresponding theoretical concepts.\\
The exclusion of Style, Setting, and Situatedness defines the scope of
the primary narrative organization analysis rather than assuming that
these dimensions have no residual convergence. Style focuses on lexical
and sentence-level attributes, whereas Setting and Situatedness may
encompass information explicitly mentioned in prompts, although this is
also possible for features in the included dimensions. The exploratory
analysis of 304 StoryScope features spanning ten dimensions examines
whether including the three excluded dimensions makes a difference to
the equally weighted mean of dimension-level RCIs (Section 3.9).

\subsection{3.3. Mixed-type narrative
distance}\label{mixed-type-narrative-distance}

Since the selected narrative features have different data types, it is
not appropriate to measure their distances using a single distance
criterion. More specifically, the set includes features with only two
values, categorical features, features with ordered values, numeric
features, and features for which several options can be selected
simultaneously. To address this problem, the study measures the
distances between stories using a method similar to Gower's (1971). For
binary and categorical features, the feature-level distance is zero when
the values are identical in both stories and one when they differ. For
numerical attributes, the distance is calculated relative to the range
of their values in the observed data and normalized to the unit
interval. For ordinal features, predominantly numeric responses are
normalized using their observed range; otherwise, answer options are
assigned normalized ranks in the order specified by the taxonomy, with
sorted observed values used as a fallback when no order is provided. The
distance is the absolute difference between the normalized values. For
attributes allowing more than one option to be selected at once, Jaccard
distance is used.

Each attribute has equal weight within its narrative dimension when the
values are known. The distance for each dimension is calculated as the
arithmetic mean of the available feature distances. If an attribute
value is unknown for either story in a comparison, its feature distance
is excluded. That is, a missing value is treated as neither agreement
nor disagreement, and no estimation is used to fill the gap. The
analysis of 51,240 stories shows that, of the 10,914,120 data cells
corresponding to stories and narrative features, 5,935 have missing
values. This represents 0.0544\% of the feature cells. Online Resource 1
provides the distribution of these missing values by individual feature,
narrative dimension, and model source. However, the distance rules for
ordinal and multi-select features, as well as the aggregation procedure,
are specific to this study and may differ from Gower's original
framework.

A lower narrative-distance value indicates greater similarity between
stories in the coded characteristics within a particular dimension.
Since raw narrative distances may not be directly comparable across
different narrative dimensions, this study relies primarily on the
proportional RCI for cross-dimensional comparisons.

\subsection{3.4. Semantically matched different-prompt
controls}\label{semantically-matched-different-prompt-controls}

The matched-prompt baseline addresses the general content of the
prompts. The prompts are represented numerically using the
sentence-transformers/all-MiniLM-L6-v2 model, and the similarity between
them is calculated using cosine similarity to identify the three most
similar but distinct prompts for each focal prompt. The similarities
between each focal prompt and its single closest distinct prompt have a
mean of .668, a median of .660, and an IQR of .631--.694. Comparison
with semantically similar but different prompts provides a more closely
matched point of comparison than random selection of prompts. Sentence
embedding comparison in this manner follows the general principle used
in Reimers and Gurevych (2019). However, their research does not
validate this particular MiniLM model.

For any model pair (A, B) and main prompt (p), the same-prompt distance
Dsame(Ap, Bp) denotes the dissimilarity between the stories produced by
two models in response to the same prompt. The matched-baseline distance
is calculated as the average distance between the story produced by
model A for the main prompt and the stories produced by model B for the
three most similar but different prompts (q). In both comparisons, model
A provides the story for the main prompt, whereas model B provides the
story or stories used for comparison. The matched-control comparison is
directional; sensitivity to reversing model roles is tested in Section
3.9. Semantic similarity between prompts does not establish task
equivalence (Roberts et al., 2020). A separate exploratory audit of
narrative-instruction alignment is described in Section 3.11.

As another way to check the stability of the results, I change the
method of finding similar prompts. While the main method uses a model
that represents the meaning of sentences numerically, in this case I use
the TF-IDF method. TF-IDF measures lexical similarity by weighting words
according to their frequency and rarity across prompts, taking into
account unigrams and bigrams. Thus, the method of calculating similarity
and finding similar prompts changes, while the narrative features and
the main semantic mask remain the same. If the results remain similar
under both methods, this reduces the possibility that the main finding
depends solely on the characteristics of the sentence-embedding model.

\subsection{3.5. Prompt-relevance masking as a sensitivity
analysis}\label{prompt-relevance-masking-as-a-sensitivity-analysis}

The primary matched-prompt baseline addresses broad similarity in prompt
content. As a complementary sensitivity analysis, a ``semantic mask'' is
implemented to check whether the results of the comparison change when
narrative features showing the highest semantic similarity with the
prompts are excluded from the distance calculations for those prompts.
For this reason, I use the descriptions of the chosen narrative features
from the StoryScope corpus, including the feature name, question, and
answer options. After that, I transform these descriptions into
numerical embeddings by using the same all-MiniLM-L6-v2 model applied to
prompts. In this way, I can estimate the semantic similarity between
each prompt and each narrative feature description. At the next step,
under the primary 99th-percentile masking condition, Iselect the top 1\%
of all prompt--feature pairs with the highest scores and exclude these
feature comparisons when calculating distances for the respective
prompts. For this purpose, the threshold is set to 0.3273. On average,
using this threshold, approximately two features are excluded from the
distance calculation out of the 213 total features for each prompt. For
matched-control comparisons, a feature is included only when it is
unmasked for both the focal prompt and its matched prompt.

RCI is similarly computed after masking at the 95th, 97.5th, and 99.5th
percentile levels to test sensitivity to the cutoff. The 95th-percentile
cutoff removes around eleven features from each prompt. However, this
masking scheme is not perfect, particularly near the classification
threshold (Section 4.1). Comparison with the main mask then tests the
stability of the results.

\subsection{3.6. Construct-validity audit and annotation
protocol}\label{construct-validity-audit-and-annotation-protocol}

In order to evaluate the validity of the semantic mask, 420
prompt--feature pairs were selected on the basis of the semantic mask
results. These 420 samples were subdivided into three categories: 140
prompt--feature pairs having a high degree of semantic similarity; 140
having a moderate degree of semantic similarity and being close to the
decision threshold of the automatic method; and 140 having a low degree
of semantic similarity. These 420 samples were associated with 396
distinct prompts. The samples were also selected so that the decisions
of the automatic method were equally distributed. This means that in 210
cases the semantic mask classified the feature as semantically relevant
to the prompt, and in the other 210 cases, it did not. Annotator A (the
author) had access to the prompt text, feature identifier, feature name,
the question used to identify the feature, and its narrative dimension.
At the same time, the annotator did not have access to the semantic
similarity score, sample category, semantic mask decision, or other
outputs of the automatic system.

Explicit cases are those in which the prompt directly requires a
specific narrative feature. Implicit cases are those in which the prompt
strongly constrains a particular feature without explicitly requiring
it. Open cases are those in which the prompt neither requires nor
strongly constrains a particular feature. For the binary evaluation,
Explicit and Implicit cases are classified as prompt-relevant, while
Open cases are classified as not prompt-relevant. Annotator A's labels
for all 420 prompt--feature combinations served as the reference values
for the semantic-mask evaluation. The author subsequently checked the
retained reference labels without making any changes. Because the sample
was stratified, the results provide insight into classification
performance on the selected sample rather than the frequency
distribution across the whole corpus. Multiple interpretations may also
be reasonable (Plank, 2022). The reporting of the coding procedure and
its evaluation draws on the guidelines of van der Lee et al. (2019).

\subsection{3.7. Residual Convergence Index and
uncertainty}\label{residual-convergence-index-and-uncertainty}

For each narrative dimension, RCI compares cross-model distance under
identical prompts with distance under distinct but semantically similar
prompts. It is calculated as follows:

\[\mathrm{RCI}=\frac{D_{\mathrm{matched}}-D_{\mathrm{same}}}{D_{\mathrm{matched}}}\]

In the current case, D\_same stands for the distance between stories
produced by two different models based on one and the same prompt.
D\_matched stands for the baseline distance calculated as the average
distance between a story produced by one model for the main prompt and
stories produced by the other model for the three most similar but
different prompts. Zero RCI indicates that the mean distance between
stories generated from the same prompt is equal to the mean distance in
the matched condition. An RCI value of .40 indicates that the mean
distance between same-prompt stories is 40\% less than the mean distance
in the baseline condition. The reverse situation is indicated by a
negative RCI. In the core analysis of 213 features across seven
narrative dimensions, the same-prompt and matched-control distances are
calculated for each of the ten model pairs and each main prompt, and
then aggregated across 10,248 prompts. RCI for each of the narrative
dimensions is calculated from the two overall mean distances, rather
than as the average of the RCIs calculated for individual prompts.
However, the 14-feature recoding analysis and the StoryScope comparisons
based on the same sample use a different aggregation approach. In these
analyses, the RCI is first calculated for each focal prompt, and then
the valid prompt-level RCIs are averaged (Sections 3.10 and 4.7).

To measure the uncertainty in the results, 95\% confidence intervals
(CIs) are calculated. To do this, I repeatedly resample the prompts
1,000 times with replacement. During each resampling process, I keep all
comparisons pertaining to a particular prompt together because these
comparisons are not independent of each other. The confidence intervals
reflect uncertainty under resampling the StoryScope prompts for the five
fixed model sources. The relatively narrow confidence intervals partly
reflect the large number of prompts analyzed. However, these confidence
intervals do not establish that the results generalize to other models,
model providers, model versions, datasets, or approaches to feature
extraction. Nonetheless, the resampling process does not account for
dependencies that may exist between different prompts, particularly when
the same story serves as a matched control for multiple prompts. Another
sensitivity test, which uses fixed matches and a multiplier based on
prompt IDs, examines the potential effects of such dependence between
prompts. This test is conducted only for the fourteen main estimators:
seven narrative dimensions under two conditions.

For the shared-ID test, each focal-prompt datum corresponds to four
prompts, which are the focal prompt plus its three matched neighbors.
Each unique prompt identifier receives its own centered and
unit-variance random multiplier, which is then reused anytime that
particular ID appears as a focal or control prompt. The contributions
from these four prompt IDs are equally weighted and divided by the
square root of four. These multipliers modify the linearized
contributions of the focal prompts to the ratio-of-mean-distances RCI.
The shared-ID standard errors are compared with ordinary focal-prompt
influence-function standard errors, and exploratory 95\% confidence
intervals are formed based on a normal-reference approximation.
Furthermore, 5,000 focal-prompt bootstraps are taken to compare the
scale of their confidence intervals with those obtained from the first
1,000 bootstraps. The neighbor graph remains the same. The diagnostic is
done only on the fourteen primary estimators and not on the TF-IDF or
caliper analyses (Online Resource 1).

\subsection{3.8. Semantic-caliper
robustness}\label{semantic-caliper-robustness}

One of the issues that remain in this research is that the closest
different prompts can still be quite dissimilar to the main prompt. This
difference can make the matched-control story distance larger than it
would be with more similar prompts. In order to study this issue, I
perform another supplementary experiment in which, unlike the primary
analysis, I select the single different prompt that has the highest
similarity with each main prompt. then gradually tighten the similarity
requirement, retaining only main prompts whose closest different prompt
meets the specified threshold. I use six thresholds for cosine
similarity: 0.55, 0.60, 0.65, 0.70, 0.75, and 0.80.

The selection criterion maintains the same 213 features, the original
semantic mask, and ten model pairings. In the matched-control
comparisons, a feature is included only if it is unmasked for both
prompts. Uncertainty is measured using prompt-level bootstrapping
(Section 3.7). The strict-caliper reporting criterion selects the
highest caliper value that maintains at least 500 focal prompts. In this
case, it selects the .75 criterion (690 prompts), while the .80
criterion maintains 383 prompts and is reported for descriptive purposes
only. Because calipers also change sample composition, these
observational comparisons do not identify causal prompt effects (Stuart,
2010).

Tighter calipers select prompts whose nearest semantic neighbors are
more similar. Consequently, the prompts retained at the .75 threshold
may differ from those in the full dataset. To examine this possibility,
I check the lengths of prompt bodies after excluding the common
instruction on the required story length. I also examine the required
story length separately. To compare variation in prompt content, I
categorize the 10,248 prompt bodies into 20 clusters using TF-IDF
unigrams and bigrams with the MiniBatchKMeans clustering technique (seed
20260826). I then measure the association between cluster membership and
inclusion in the strict .75 subset using Cramér\textquotesingle s V.
This analysis is descriptive and does not change the weights assigned to
prompts or the RCI estimates.

\subsection{3.9. Feature-family, aggregation-weighting, and model-role
sensitivity}\label{feature-family-aggregation-weighting-and-model-role-sensitivity}

The Feature Family Sensitivity Test is conducted to investigate whether
the overall result depends on a particular narrative dimension or on
giving equal weight to all seven narrative dimensions. The first test
compares the equally weighted mean of the seven dimension-level RCIs
with a feature-count-weighted mean, in which dimensions containing more
features receive greater weight. The equally weighted mean is then
calculated seven times, leaving out one dimension each time. These tests
change how results from the narrative dimensions are combined or which
dimensions are included, but the extracted feature values remain
unchanged. All weighting tests use the narrative feature values
extracted by StoryScope, whereas the separate recoding test described in
Section 3.10 uses author-reviewed recoded feature values.

I also perform the comparison using the three nearest distinct prompts
across all ten StoryScope dimensions (304 features) for all 10,248 focal
prompts. In particular, RCI for each dimension is estimated as the ratio
of mean distances. The seven-dimension and ten-dimension summaries are
calculated by averaging these dimension-level RCIs with equal weights,
rather than pooling all 304 features together. The three additional
dimensions are Style (39 features), Setting (27), and Situatedness (25).
The masking process maintains the similarity threshold of about 0.3273,
originally set at the 99th percentile of the seven-dimension
distribution, rather than recalculating it using the 304-feature
distribution. Paired focal-prompt-level bootstrap resampling (4,000
replicates; seed 20261020) is used to estimate uncertainty in the
dimension-level RCIs and their equally weighted averages under both
masked and unmasked conditions. Percentile 95\% confidence intervals are
constructed for the paired differences between the ten-dimension and
seven-dimension RCI summaries.

A further sensitivity analysis examines whether reversing the roles of
models A and B affects the RCI estimates calculated using the 213
narrative features. For each unordered model pair (A, B), each original
prompt p, and each of its three matched prompts q, the original distance
d(Ap, Bq) is compared with the reversed distance d(Bp, Aq). The
symmetric matched distance is the arithmetic mean of these two
distances. The same-prompt distance, d(Ap, Bp), remains unchanged. Only
comparisons valid in both directions are included when calculating the
symmetric matched distance. The RCI for each dimension is then
calculated from the ratio of mean distances, and the seven
dimension-level RCIs are averaged with equal weights. For both unmasked
and masked 99th-percentile comparisons, the entire set of 10,248 focal
prompts is used together with the original fixed three-neighbor mapping.
The procedure first reproduces the original same-prompt and forward
matched-control distances for the fourteen primary dimension-condition
combinations to within 10\ensuremath{^{-6}}. Paired focal-prompt percentile confidence
intervals are then calculated from 4,000 bootstrap samples (seed
20261022). The matching graph remains fixed throughout bootstrapping,
which does not fully account for dependence arising from reused matched
controls. All output data are available in Online Resource 1 under
Symmetric\_213\_Feature\_Model\_Role\_Sensitivity/.

\subsection{3.10. Story-level feature recoding
sensitivity}\label{story-level-feature-recoding-sensitivity}

The story-level recoding analysis examines whether the results are
sensitive to the feature values extracted by StoryScope. Before
examining the feature values, 14 features were selected. Two features
were selected from each narrative dimension, representing different
response types. These response types included categorical, binary,
ordinal, numeric, and multi-select. A total of 40 main prompts were
selected. Of these, 20 had a cosine similarity of at least .75 with
their nearest distinct prompt, while the other 20 were randomly selected
from the full dataset. For each of the 40 main prompts, the five
model-generated stories were coded, giving a total of 200 stories. Next,
the nearest distinct prompt was selected as a matched control for each
of the 40 main prompts. These control prompts were different from the
main prompts selected earlier and contributed another 200 story records,
bringing the total to 400.

The feature values were assigned based on the contents of the stories,
rather than their relevance to the prompts. The author, acting as
Annotator A, had access to the full texts of the stories and Story\_IDs.
The annotator did not have access to the prompts, identities of the
models producing the stories, their focal or control status, semantic
similarity scores, or original feature values from StoryScope. The
author reviewed preliminary AI-assisted labels assigned from story
content, without information about model identities or story status, and
subsequently reviewed the retained labels for all 400 stories.

The retained reference labels form the basis of the recoding analysis
and the direct comparison with StoryScope. Previous research
demonstrates that large language models can assist with some annotation
tasks (Gilardi et al., 2023). Nonetheless, this does not validate the
specific approach employed here.

For a model pair (A, B), the matched-control distance (Dmatched) is
calculated in only one direction, meaning that the story produced by
model A in response to the main prompt (p) is compared with the story
produced by model B in response to the matched prompt (q). The distances
for the 14 features are calculated between the two stories and then
aggregated to obtain a story-pair distance. For each main prompt, the
distances are averaged across the ten model pairs before calculating its
RCI.

The overall 14-feature RCI is computed by aggregating distances across
all 14 recoded features, not by averaging the RCI scores for each of the
seven narrative dimensions. The valid prompt-level RCIs are then
averaged. Each narrative dimension is also individually considered based
on its selected features. When the matched-control distance for a prompt
is zero, the RCI score for that narrative dimension is not computed.
These cases are excluded from the computation of the average score for
that dimension. Therefore, the Agents dimension contains 36 prompts with
valid RCI scores, whereas the Temporal Structure dimension contains 35
such prompts. All 40 prompts, however, contribute valid RCI scores to
the overall 14-feature analysis. The 95\% confidence intervals are
computed through 10,000 bootstrap samples of the main prompts (seed
20261009). During resampling, all model-pair comparisons associated with
a prompt are kept together.

The sensitivity analysis also employs symmetric comparisons.
Specifically, the distance between the story produced by model A in
response to the main prompt and the story produced by model B in
response to the matched prompt is averaged with the reversed distance,
in which model B provides the main-prompt story and model A provides the
matched-prompt story. No prompt-relevance masking is employed when
measuring distances using the 14 recoded features because the purpose of
this additional analysis is to check the sensitivity of the results to
feature values, not to replace the main analysis involving 213 features.
For the direct comparison with StoryScope, the same prompts, features,
matching rule, and aggregation procedure are used; only the feature
values differ. I compute the RCI using both sets of values. Uncertainty
is measured using 10,000 paired bootstrap resamples of the prompts.

\subsection{3.11. Exploratory audit of narrative-constraint
alignment}\label{exploratory-audit-of-narrative-constraint-alignment}

Exploratory analyses of prompt constraints were conducted based on 120
directed pairs of focal--control items corresponding to 55 focal prompts
within an archival pilot sample. The selection was non-random. Each pair
was rated according to six aspects: narrative perspective, characters
and roles, required central events, conflict, relationships, and
narrative development or ending. For each aspect, the requirements were
evaluated in terms of conflict, partial specification, agreement, and
mutual lack of specification. I then assigned an overall grade
reflecting very little overlap (0), thematic or contextual similarity
(1), substantial but incomplete agreement (2), and near-equivalence (3).
Based on the existing masked narrative distances, I then compared RCI
values between more-aligned (grades 2--3) and less-aligned (grades 0--1)
controls for the same focal prompts. Secondary rank- and
similarity-adjusted measures at the link level examined potential
confounding. However, these distance-based statistics do not represent
adjusted RCIs (see Online Resource 1).

A second sample, comprising 120 focal prompts and their three closest
distinct neighboring prompts (360 directed dyads), was also generated
using lexical stratification to augment the audit. The same
six-criterion AI-assisted rubric was employed again. The normalized
embeddings for all-MiniLM-L6-v2 have been recalculated using the
archived prompts to verify the neighbor mapping. The cosine similarity
scores have been computed for all the 30,744 pairs, and the deviation
from all 10,248 archived nearest-neighbor scores is less than 5.36 \ensuremath{\times}
10\ensuremath{^{-7}}.There were no changes in the neighbor ranking order. The masked
distances have been calculated for 360 pairs for 10 model pairs and 7
dimensions; 4 distances could not be imputed.RCI comparisons within
focal prompts used only focal prompts with controls in both alignment
groups. Fixed-effects diagnostics separately adjusted mean
matched-minus-same narrative distance for verified MiniLM cosine, with a
close-cosine sensitivity test. Uncertainty was measured using 4,000
focal-prompt bootstrap samples. Sampling details, model specifications,
and additional findings have been preserved in Online Resource 1.

\subsection{3.12. Reproducibility, and generation
metadata}\label{reproducibility-and-generation-metadata}

For this analysis, the feature values have been obtained from
storyscope\_features.parquet from the StoryScope dataset. Full
information on the text-generation parameters, such as temperature,
top-p, and random seed values, is not available for certain models.
Here, neither regeneration of the stories nor re-running of the
StoryScope feature extraction procedure was done. The code repositories
and dataset-card version are mentioned in the Data Availability section.
The code for the analysis, along with file manifests, validation audits,
prompt-neighbor mappings, robustness analysis output, and additional
export files, has been provided in Online Resource 1 (ESM\_1.zip). The
MiniBatchKMeans model was fitted using random seed 20260826.
Bootstrapping procedures with their respective random seed values are
explained alongside each analysis. The checksum of the current archive
is provided in the Code and Materials Availability section.

\section{4. Evaluation results}\label{evaluation-results}

\subsection{4.1. Construct-validity results for the prompt-relevance
mask}\label{construct-validity-results-for-the-prompt-relevance-mask}

Table 2 compares the automatic semantic-mask decisions with the
author\textquotesingle s reference labels (Annotator A) for 420
prompt--feature pairs, classified as Explicit, Implicit, or Open.
Explicit and Implicit are combined into the Prompt-relevant category for
binary classification. Compared with these reference labels, the
automatic mask demonstrated an accuracy of .819 and an F1 score of .813.
However, for cases close to the classification threshold, accuracy and
F1 were considerably lower, at .621 and .644, respectively.

Of the 76 disagreements between the mask and the reference labels, 53
occurred near the automatic cutoff. Accuracy for the high-similarity
group was .836, and for the low-similarity group it was 1.000. However,
because the sample is stratified and balanced according to the mask
classifications, I cannot use these figures to estimate accuracy for the
entire dataset. The concentration of disagreements near the threshold
indicates that the semantic mask does not consistently identify
prompt-relevant features in these cases.

\begin{longtable}[]{@{}
  >{\raggedright\arraybackslash}p{(\columnwidth - 2\tabcolsep) * \real{0.46}}
  >{\raggedright\arraybackslash}p{(\columnwidth - 2\tabcolsep) * \real{0.54}}@{}}
\caption{Construct-validity audit of the semantic prompt-relevance mask}\label{tab:2}\\
\toprule\noalign{}
\begin{minipage}[b]{\linewidth}\raggedright
\textbf{Measure}
\end{minipage} & \begin{minipage}[b]{\linewidth}\raggedright
\textbf{Result}
\end{minipage} \\
\midrule\noalign{}
\endfirsthead
\toprule\noalign{}
\begin{minipage}[b]{\linewidth}\raggedright
\textbf{Measure}
\end{minipage} & \begin{minipage}[b]{\linewidth}\raggedright
\textbf{Result}
\end{minipage} \\
\midrule\noalign{}
\endhead
\bottomrule\noalign{}
\endlastfoot
Validation sample & 420 cells: 140 high, 140 near-threshold, 140 low \\
Final three-way labels & 58 Explicit; 138 Implicit; 224 Open \\
Binary confusion counts & TP 165; TN 179; FP 45; FN 31 \\
Accuracy / balanced accuracy & .819 / .820 \\
Precision / recall & .786 / .842 \\
Specificity / F1 & .799 / .813 \\
Mask vs final-label kappa / Matthews correlation coefficient & .638 /
.640 \\
Near-threshold accuracy / F1 & .621 / .644 \\
Reference-label basis & Annotator A labels retained without change \\
\end{longtable}

Note: Explicit and Implicit count as prompt-relevant; Open is
human-open. The balanced, stratified audit is not a prevalence estimate
for the corpus.

\subsection{4.2. Primary matched-control convergence and mask
sensitivity}\label{primary-matched-control-convergence-and-mask-sensitivity}

The primary analysis based on the matched-control technique shows that
stories created by different models in response to the same prompt are
more similar in their coded narrative features than stories generated
from semantically similar but different prompts, even before applying
the semantic mask. Across the seven narrative dimensions, the values of
the Residual Convergence Index (RCI) range from .326 for Temporal
Structure to .526 for Perspective, and their average is .409. After
masking at the 99th-percentile threshold, the RCI values range from .323
to .508, and their average is .406. This represents a decrease of
approximately .004 using the unrounded estimates. Consequently, masking
has only a small effect on the average RCI in this analysis. The RCI
values remain positive in all seven narrative dimensions after masking
at the 99th-percentile threshold (Table 3). Nevertheless, the
differences among these dimensions are descriptive and should not be
treated as a definitive ranking of the narrative dimensions.

Another sensitivity analysis considers how the uncertainty estimates may
be affected by dependencies among matched prompts. This sensitivity
analysis uses the observed matching pattern, which includes 30,744
prompt matches involving 10,248 main prompts and 7,617 unique control
prompts, according to the three-nearest-neighbor rule. The shared-ID
sensitivity procedure produces standard errors 9--17\% larger across the
fourteen combinations of narrative dimensions and masking conditions.
Nevertheless, the lower bounds of the exploratory 95\% confidence
intervals remain positive for all fourteen estimates. The sensitivity
analysis is exploratory: it holds the matching graph fixed, gives equal
weight to the focal and three neighboring IDs, and reuses each
ID-specific multiplier wherever that prompt occurs. It does not estimate
the true correlations among prompts. Thus, all the main tables and
figures continue to use the original confidence intervals obtained by
bootstrapping the main prompts (Online Resource 1).

\begin{figure}[htbp]
\centering
\includegraphics[width=.90\linewidth]{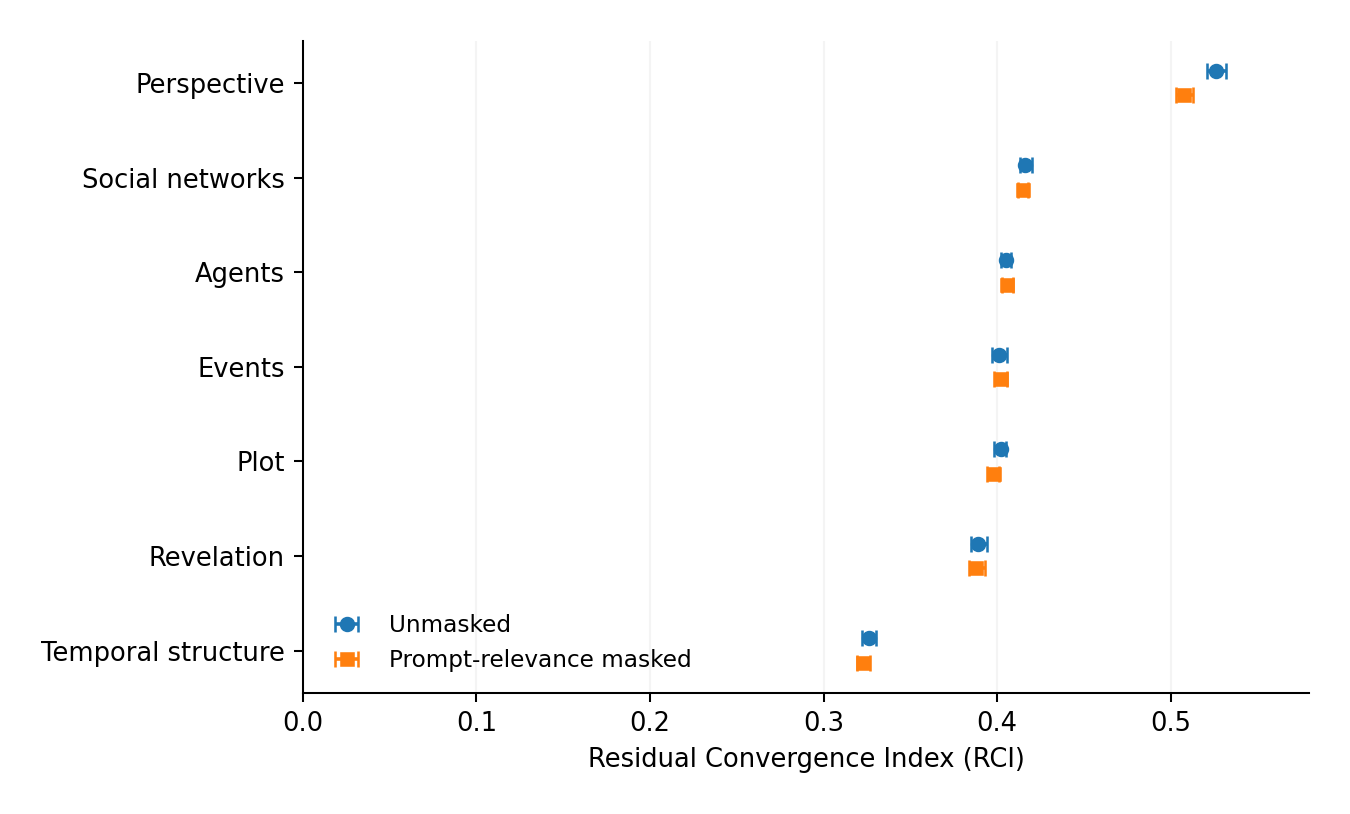}
\caption{Residual Convergence Index for each narrative dimension pre- and post-masking of prompt-relevance. Error bars represent 95\% confidence intervals on the prompt bootstraps. The near-overlap of the estimates suggests that masking the strongest prompt-feature alignment makes only a small difference to the overall outcome.}\label{fig:1}
\end{figure}

\begin{longtable}[]{@{}
  >{\raggedright\arraybackslash}p{(\columnwidth - 8\tabcolsep) * \real{0.19}}
  >{\raggedright\arraybackslash}p{(\columnwidth - 8\tabcolsep) * \real{0.09}}
  >{\raggedright\arraybackslash}p{(\columnwidth - 8\tabcolsep) * \real{0.24}}
  >{\raggedright\arraybackslash}p{(\columnwidth - 8\tabcolsep) * \real{0.24}}
  >{\raggedright\arraybackslash}p{(\columnwidth - 8\tabcolsep) * \real{0.24}}@{}}
\caption{Primary and alternative prompt-matching estimates}\label{tab:3}\\
\toprule\noalign{}
\begin{minipage}[b]{\linewidth}\raggedright
\textbf{Dimension}
\end{minipage} & \begin{minipage}[b]{\linewidth}\raggedright
\textbf{Features}
\end{minipage} & \begin{minipage}[b]{\linewidth}\raggedright
\textbf{Unmasked RCI {[}95\% CI{]}}
\end{minipage} & \begin{minipage}[b]{\linewidth}\raggedright
\textbf{Masked RCI {[}95\% CI{]}}
\end{minipage} & \begin{minipage}[b]{\linewidth}\raggedright
\textbf{TF-IDF matched + mask RCI {[}95\% CI{]}}
\end{minipage} \\
\midrule\noalign{}
\endfirsthead
\toprule\noalign{}
\begin{minipage}[b]{\linewidth}\raggedright
\textbf{Dimension}
\end{minipage} & \begin{minipage}[b]{\linewidth}\raggedright
\textbf{Features}
\end{minipage} & \begin{minipage}[b]{\linewidth}\raggedright
\textbf{Unmasked RCI {[}95\% CI{]}}
\end{minipage} & \begin{minipage}[b]{\linewidth}\raggedright
\textbf{Masked RCI {[}95\% CI{]}}
\end{minipage} & \begin{minipage}[b]{\linewidth}\raggedright
\textbf{TF-IDF matched + mask RCI {[}95\% CI{]}}
\end{minipage} \\
\midrule\noalign{}
\endhead
\bottomrule\noalign{}
\endlastfoot
Perspective & 17 & 0.526 {[}0.521, 0.532{]} & 0.508 {[}0.503, 0.513{]} &
0.479 {[}0.474, 0.485{]} \\
Social networks & 39 & 0.416 {[}0.413, 0.420{]} & 0.415 {[}0.412,
0.418{]} & 0.434 {[}0.431, 0.437{]} \\
Agents & 54 & 0.405 {[}0.402, 0.408{]} & 0.406 {[}0.403, 0.409{]} &
0.420 {[}0.417, 0.423{]} \\
Events & 26 & 0.401 {[}0.397, 0.406{]} & 0.402 {[}0.398, 0.406{]} &
0.408 {[}0.404, 0.411{]} \\
Plot & 28 & 0.402 {[}0.398, 0.405{]} & 0.398 {[}0.394, 0.401{]} & 0.411
{[}0.408, 0.414{]} \\
Revelation & 25 & 0.389 {[}0.385, 0.394{]} & 0.388 {[}0.384, 0.393{]} &
0.394 {[}0.389, 0.398{]} \\
Temporal structure & 24 & 0.326 {[}0.322, 0.330{]} & 0.323 {[}0.319,
0.327{]} & 0.328 {[}0.323, 0.332{]} \\
\end{longtable}

Note: All estimates use 10,248 prompts. TF-IDF rebuilds the
three-nearest-distinct-prompt baseline using unigrams and bigrams,
retaining the primary semantic mask.

\subsection{4.3. Mask-threshold
sensitivity}\label{mask-threshold-sensitivity}

The sensitivity analysis results using the semantic mask are not heavily
dependent on the 99th-percentile cutoff. With the more extensive mask at
the 95th-percentile cutoff, the mean RCI across dimensions becomes .402
rather than .406. The 97.5th-, 99th-, and 99.5th-percentile cutoffs
produce mean RCI values of .403, .406, and .407, respectively. Thus, the
result changes little when approximately eleven features per prompt are
excluded instead of approximately two.

\subsection{4.4. Alternative prompt
matching}\label{alternative-prompt-matching}

Using unigram/bigram TF-IDF instead of MiniLM to identify matched
controls from distinct prompts, the mean masked RCI is .410, compared
with .406 under MiniLM matching (Table 3). While the RCI for the
Perspective dimension decreases, those for the other dimensions increase
slightly.

\subsection{4.5. Semantic-caliper
robustness}\label{semantic-caliper-robustness-1}

When I calculate the matched-control baseline using only the closest
distinct prompt, the mean RCI across the seven dimensions in the full
sample is .397. This value is slightly lower than the mean of .406
obtained in the main masked analysis, which uses three similar prompts.
In the single-neighbor approach, RCI ranges from .317 (Temporal
Structure) to .499 (Perspective). Within this single-neighbor analysis,
the mean RCI generally decreases as the minimum similarity threshold
becomes stricter. At a similarity threshold of .65 or above, 6,048
prompts are retained, with a mean RCI of .389. At a threshold of .70 or
above, 2,223 prompts are retained, and the mean RCI decreases to .352.

These stricter similarity criteria also change the composition of the
selected subset. The most stringent threshold retaining at least 500
main prompts was .75, resulting in a total of 690 prompts. The mean
similarity between each retained main prompt and its nearest distinct
prompt is .821. The lower bounds of the 95\% confidence intervals are
positive for all seven narrative dimensions, and the mean RCI across
dimensions is .250 (see Table 4).

\begin{longtable}[]{@{}
  >{\raggedright\arraybackslash}p{(\columnwidth - 4\tabcolsep) * \real{0.37}}
  >{\raggedright\arraybackslash}p{(\columnwidth - 4\tabcolsep) * \real{0.2}}
  >{\raggedright\arraybackslash}p{(\columnwidth - 4\tabcolsep) * \real{0.43}}@{}}
\caption{Residual convergence under the strict semantic caliper (nearest distinct prompt cosine similarity \textgreater= .75; n = 690; mean similarity = .821)}\label{tab:4}\\
\toprule\noalign{}
\begin{minipage}[b]{\linewidth}\raggedright
\textbf{Dimension}
\end{minipage} & \begin{minipage}[b]{\linewidth}\raggedright
\textbf{RCI}
\end{minipage} & \begin{minipage}[b]{\linewidth}\raggedright
\textbf{95\% CI}
\end{minipage} \\
\midrule\noalign{}
\endfirsthead
\toprule\noalign{}
\begin{minipage}[b]{\linewidth}\raggedright
\textbf{Dimension}
\end{minipage} & \begin{minipage}[b]{\linewidth}\raggedright
\textbf{RCI}
\end{minipage} & \begin{minipage}[b]{\linewidth}\raggedright
\textbf{95\% CI}
\end{minipage} \\
\midrule\noalign{}
\endhead
\bottomrule\noalign{}
\endlastfoot
Perspective & .314 & {[}.283, .343{]} \\
Social networks & .257 & {[}.240, .274{]} \\
Plot & .253 & {[}.234, .272{]} \\
Agents & .250 & {[}.232, .265{]} \\
Revelation & .246 & {[}.225, .270{]} \\
Events & .246 & {[}.227, .265{]} \\
Temporal structure & .184 & {[}.163, .204{]} \\
\end{longtable}

Of the 10,248 prompts, 690 (6.7\%) have a closest distinct neighbor with
a cosine similarity of at least .75. The average prompt-body length is
nearly identical for the .75 subset and the remaining prompts (83.6
versus 83.7 words; rank-biserial correlation = -.007). The median
requested story length is higher in the .75 subset (6,000 versus 5,000
words), although the rank-biserial correlation is small (.081). All 20
TF-IDF prompt clusters are represented in the .75 subset, although their
proportions vary (Cramér\textquotesingle s V = .159).

With the more restrictive similarity threshold of .80, 383 prompts are
retained, and the RCI estimates and the lower bounds of their 95\%
confidence intervals remain positive in all seven dimensions. The mean
RCI decreases to .174, ranging from .122 for Temporal Structure to .235
for Perspective. Since the number of prompts in this subset is below the
minimum of 500 main prompts required by the reporting rule, the .80
results are treated as supplementary sensitivity results rather than the
primary strict-caliper result. I do not extend the RCI curve beyond .80.
Both the similarity of the matches and the composition of the selected
prompts change as the threshold increases. Therefore, whether RCI
approaches or crosses zero at higher similarity thresholds remains
unclear, and extrapolating the observed decline would depend heavily on
assumptions about this relationship and the changing sample composition.
Addressing this question would require a larger sample of prompts with
highly similar distinct neighbors.

In addition, an exploratory author audit of 120 pairs of prompts (55
focal prompts) found that two pairs were rated as nearly equivalent,
while 19 were rated as being substantially but not completely aligned in
terms of narrative instructions. In 15 focal prompts for which there
were controls in both groups, the average RCI of controls that were more
aligned was .289, while the average RCI of controls that were less
aligned was .414. Moreover, these groups were different in neighbor rank
and semantic similarity. Once corrected for the MiniLM cosine
similarity, the within-focal correlation between mean masked
matched-minus-same narrative distance was \ensuremath{-}.051 (.051).

Of the 360 pair comparisons (120 focal prompts), two pairs were found to
be nearly equivalent, whereas 95 pairs displayed partial but incomplete
alignment. The mean MiniLM cosine of 97 pairs with greater alignment was
.802, as compared to .681 for 263 pairs with lesser alignment. The mean
masked RCIs of 16 focal prompts with both control groups were .285 and
.337, respectively. The adjusted cosine narrative distance metric of
matched--same pairs was \ensuremath{-}.0037 (95\% bootstrap CI {[}\ensuremath{-}.0222, .0170{]}).
Thus, the extended audit does not support a stable alignment-specific
effect separate from embedding similarity. Specifications and prior
rankings can be seen in Online Resource 1.

\subsection{4.6. Feature-family, aggregation-weighting, and model-role
sensitivity}\label{feature-family-aggregation-weighting-and-model-role-sensitivity-1}

Alternative weighting and leave-one-dimension-out sensitivity analyses
yield similarly positive estimates (Table 5). The results from the main
masked analysis show that the equally weighted mean RCI is .406, while
the feature-count-weighted mean is .403; the leave-one-dimension-out
estimates range from .389 to .420. With the .75 semantic caliper
threshold, the corresponding means are .250 and .248, while the
leave-one-dimension-out estimates range from .239 to .261. Although
these analyses suggest that the positive estimates are not strongly
dependent on the choice of dimensions or weighting method, they do not
independently validate the feature values.

\begin{longtable}[]{@{}
  >{\raggedright\arraybackslash}p{(\columnwidth - 6\tabcolsep) * \real{0.29}}
  >{\raggedright\arraybackslash}p{(\columnwidth - 6\tabcolsep) * \real{0.23}}
  >{\raggedright\arraybackslash}p{(\columnwidth - 6\tabcolsep) * \real{0.25}}
  >{\raggedright\arraybackslash}p{(\columnwidth - 6\tabcolsep) * \real{0.23}}@{}}
\caption{Feature-family and aggregation-weighting sensitivity}\label{tab:5}\\
\toprule\noalign{}
\begin{minipage}[b]{\linewidth}\raggedright
\textbf{Analysis regime}
\end{minipage} & \begin{minipage}[b]{\linewidth}\raggedright
\textbf{Equal-dimension mean RCI}
\end{minipage} & \begin{minipage}[b]{\linewidth}\raggedright
\textbf{Feature-count-weighted mean RCI}
\end{minipage} & \begin{minipage}[b]{\linewidth}\raggedright
\textbf{Leave-one-dimension-out range}
\end{minipage} \\
\midrule\noalign{}
\endfirsthead
\toprule\noalign{}
\begin{minipage}[b]{\linewidth}\raggedright
\textbf{Analysis regime}
\end{minipage} & \begin{minipage}[b]{\linewidth}\raggedright
\textbf{Equal-dimension mean RCI}
\end{minipage} & \begin{minipage}[b]{\linewidth}\raggedright
\textbf{Feature-count-weighted mean RCI}
\end{minipage} & \begin{minipage}[b]{\linewidth}\raggedright
\textbf{Leave-one-dimension-out range}
\end{minipage} \\
\midrule\noalign{}
\endhead
\bottomrule\noalign{}
\endlastfoot
Primary matched + 99th-percentile mask & .406 & .403 & .389--.420 \\
Strict semantic caliper (\textgreater= .75) & .250 & .248 &
.239--.261 \\
\end{longtable}

Note: Weights use 17, 39, 54, 26, 28, 25, and 24 features across the
seven dimensions.

The equally weighted masked mean value of RCI for the ten-dimension
sensitivity analysis was 0.3959 for all ten dimensions and 0.4056 for
the seven main dimensions (difference = \ensuremath{-}0.0098; paired-bootstrap 95\%
confidence interval {[}\ensuremath{-}0.0105, \ensuremath{-}0.0091{]}). Unmasked values yielded
means of 0.3984 and 0.4093 (difference = \ensuremath{-}0.0109; 95\% confidence
interval {[}\ensuremath{-}0.0117, \ensuremath{-}0.0102{]}), respectively. These calculations were
done using the same 10,248 focal prompts, three-nearest-neighbor
matching, and masking criterion described in Section 3.9. In the masked
calculation, the Setting dimension had an RCI value of 0.4491, which was
the second highest out of ten dimensions. Overall, adding these three
additional dimensions decreases the equally weighted mean value
slightly, but it remains positive. This difference changes the set of
analyzed features without establishing prompt-independent convergence.
Masked and unmasked results for all ten dimensions, paired-bootstrap
estimates, and resampling records are available in Online Resource 1.

The symmetrically averaged masked RCI for the role-reversal analysis
using all 213 features is 0.40702 (95\% paired-bootstrap confidence
interval {[}0.40449, 0.40944{]}), compared to 0.40566 (95\% confidence
interval {[}0.40310, 0.40816{]}), which is computed using the original
model roles. The difference between these two is +0.00136 (95\%
confidence interval {[}+0.00096, +0.00175{]}). For the unmasked case,
the difference between symmetric and forward averaging values is
+0.00132 (95\% confidence interval {[}+0.00093, +0.00170{]}). The
symmetric and forward averages are 0.41068 and 0.40936, respectively.
All seven dimensions provide positive estimates for symmetric roles
under both conditions. The largest masked difference between directional
and symmetric estimates is obtained for the Events dimension (+0.00346).
The valid pair-support ratio is greater than 99.7\% in all dimensions.
The results reproduce the original fourteen same-prompt and
forward-distance estimates with a maximum numerical difference of less
than 10\ensuremath{^{-6}}. This confirms that the positive matched-control contrast
still holds true when the models\textquotesingle{} roles are
symmetrized, even though there may be some discrepancies between the two
kinds of estimates. The results, together with 4,000 bootstrap
replications, are shown in Online Resource 1 under
Symmetric\_213\_Feature\_Model\_Role\_Sensitivity/. Since the symmetric
analysis includes only data that are valid in both directions, its
forward-only result can be somewhat different from the seven-dimension
estimate in the separate ten-dimension analysis.

\subsection{4.7. Author-reviewed recoding and direct measurement
sensitivity}\label{author-reviewed-recoding-and-direct-measurement-sensitivity}

Using the author\textquotesingle s recoded feature values for the sample
of 40 focal prompts, the mean Dsame is .320, while the mean Dmatched is
.404. However, the RCI is first calculated separately for each prompt,
and only afterwards are these individual values averaged. As a result,
the reported overall RCI cannot be calculated directly from the two mean
distances above. Unlike the primary RCI based on a ratio of overall mean
distances, the recoding RCI is an average of valid prompt-level ratios
and covers a smaller sample and feature set. The full-corpus .406 and
the recoding .187 are therefore not directly comparable as estimates
differing only in feature values. The overall mean RCI is .187, with a
95\% confidence interval of {[}.137, .239{]}. When the matched
comparison is performed in both directions, the mean RCI is .186, with a
95\% confidence interval of {[}.137, .235{]}. The mean RCI for the 20
randomly selected prompts is .237, with a 95\% confidence interval of
{[}.148, .318{]}, while the mean RCI for the 20 prompts selected for
their high similarity to their nearest neighbors is lower, at .138, with
a 95\% confidence interval of {[}.090, .187{]}. These subgroup
comparisons are descriptive. The lower RCI observed among the more
closely matched prompts is consistent with the general pattern found in
the similarity-threshold analysis.

When the recoded features are analyzed separately for each narrative
dimension, the positive estimates are less consistent across dimensions
than in the main analysis using StoryScope's feature values. The RCI for
Perspective is .320, with a 95\% confidence interval of {[}.186,
.452{]}; the RCI for Social Networks is .189, with a 95\% confidence
interval of {[}.097, .282{]}. The lower bounds of both confidence
intervals are positive. On the other hand, the confidence intervals for
the other five narrative dimensions include zero (Table 6). For four
main prompts in the Agents analysis and five in the Temporal Structure
analysis, Dmatched = 0. The RCI cannot be computed in these cases
because its denominator is zero. These undefined prompt-level RCIs are
excluded from the corresponding averages rather than being replaced with
zero or infinity. This applies regardless of whether the distance
between two stories generated from the same prompt is zero or positive.
The overall mean RCI remains positive in this audited sample of 40
prompts, although the dimension-level results are less consistent.
Therefore, one cannot conclude that all recoded narrative dimensions
exhibit convergence to the same extent.

\begin{longtable}[]{@{}
  >{\raggedright\arraybackslash}p{(\columnwidth - 10\tabcolsep) * \real{0.23}}
  >{\raggedright\arraybackslash}p{(\columnwidth - 10\tabcolsep) * \real{0.13}}
  >{\raggedright\arraybackslash}p{(\columnwidth - 10\tabcolsep) * \real{0.15}}
  >{\raggedright\arraybackslash}p{(\columnwidth - 10\tabcolsep) * \real{0.16}}
  >{\raggedright\arraybackslash}p{(\columnwidth - 10\tabcolsep) * \real{0.16}}
  >{\raggedright\arraybackslash}p{(\columnwidth - 10\tabcolsep) * \real{0.17}}@{}}
\caption{Annotator-A-reviewed single-nearest RCI sensitivity across 14 recoded narrative features}\label{tab:6}\\
\toprule\noalign{}
\begin{minipage}[b]{\linewidth}\raggedright
\textbf{Dimension}
\end{minipage} & \begin{minipage}[b]{\linewidth}\raggedright
\textbf{Effective n (RCI)}
\end{minipage} & \begin{minipage}[b]{\linewidth}\raggedright
\textbf{Mean Dsame}
\end{minipage} & \begin{minipage}[b]{\linewidth}\raggedright
\textbf{Mean Dmatched}
\end{minipage} & \begin{minipage}[b]{\linewidth}\raggedright
\textbf{Mean prompt-level RCI}
\end{minipage} & \begin{minipage}[b]{\linewidth}\raggedright
\textbf{95\% CI}
\end{minipage} \\
\midrule\noalign{}
\endfirsthead
\toprule\noalign{}
\begin{minipage}[b]{\linewidth}\raggedright
\textbf{Dimension}
\end{minipage} & \begin{minipage}[b]{\linewidth}\raggedright
\textbf{Effective n (RCI)}
\end{minipage} & \begin{minipage}[b]{\linewidth}\raggedright
\textbf{Mean Dsame}
\end{minipage} & \begin{minipage}[b]{\linewidth}\raggedright
\textbf{Mean Dmatched}
\end{minipage} & \begin{minipage}[b]{\linewidth}\raggedright
\textbf{Mean prompt-level RCI}
\end{minipage} & \begin{minipage}[b]{\linewidth}\raggedright
\textbf{95\% CI}
\end{minipage} \\
\midrule\noalign{}
\endhead
\bottomrule\noalign{}
\endlastfoot
Overall & 40 & .320 & .404 & .187 & {[}.137, .239{]} \\
Perspective & 40 & .315 & .521 & .320 & {[}.186, .452{]} \\
Social networks & 40 & .418 & .545 & .189 & {[}.097, .282{]} \\
Agents & 36 & .218 & .351 & .218 & {[}-.012, .422{]} \\
Events & 40 & .465 & .459 & -.008 & {[}-.089, .073{]} \\
Plot & 40 & .299 & .329 & .100 & {[}-.009, .215{]} \\
Revelation & 40 & .268 & .270 & -.061 & {[}-.181, .052{]} \\
Temporal structure & 35 & .255 & .351 & .197 & {[}-.066, .413{]} \\
\end{longtable}

\textbf{Note:} Using the author\textquotesingle s reviewed feature
values, the analysis includes 40 main prompts (20 selected randomly and
20 selected using the similarity caliper) and their 40 nearest distinct
control prompts, resulting in a total of 400 story records. The Dsame
and Dmatched values reported in each row above are averaged across all
40 main prompts. However, the sample size (n) used to calculate the mean
RCI includes only those prompts for which the RCI can be calculated.
There are 36 valid prompts for Agents, 35 for Temporal Structure, and 40
for each of the remaining dimensions. The overall RCI is calculated by
first aggregating the feature-level distances, rather than averaging the
RCI values of individual narrative dimensions. For any narrative
dimension, the prompt-level RCI is undefined when Dmatched = 0. The
bootstrap analysis uses 10,000 prompt-level resamples (seed 20261009).

In this direct same-sample comparison of measurement methods, the
following components remain constant: 40 main prompts, 40 control
prompts selected as their nearest semantic neighbors, 14 features, model
pairs, the method for computing distances between different types of
features, and the method for aggregating the results. Only the input
feature values change; at the dimension level, the number of valid
prompt-level RCIs can consequently differ, although all 40 overall
prompt-level RCIs remain defined under both coding sets. When StoryScope
feature values are used, the mean distance between stories generated
from the same prompt is .138, while using the author-reviewed coding, it
increases to .320. The mean matched-control distance is .253 when
StoryScope feature values are used, compared with .404 when the
author-reviewed coding is used.

The mean prompt-level RCI calculated using StoryScope feature values is
.380, with a 95\% confidence interval of {[}.259, .486{]}. When the
author\textquotesingle s reviewed coding is used, the mean RCI decreases
to .187, with a 95\% confidence interval of {[}.137, .239{]}. The
difference between these two estimates, calculated from the unrounded
values, is .192, with a 95\% confidence interval of {[}.081, .293{]}.
These results show that residual convergence remains positive under both
measurement approaches, but its estimated magnitude differs
substantially. The results for individual narrative dimensions and their
effective sample sizes are presented in Online Resource 1.

To examine which feature assignments differ, I compared the retained
author-reviewed and StoryScope values separately for the fourteen
selected features in the 200 focal-story records for which a historical
item-level comparison is available (Table 7). Exact agreement varies
from 2.5\% for the multi-select range of relationship types to 85.5\%
for dominant narrative person. Because whole-set agreement is stringent
for multi-select features, the mean Jaccard overlap for relationship
types is also reported (.398). Agreement is lower for the three ordinal
features collectively (37.0\%) than for the two binary features
(69.2\%), but these percentages are not directly comparable measures of
annotation quality across response types. The three areas where
disagreements occur most often are the initiative versus reactive nature
of the protagonist\textquotesingle s behavior, the explicitness of
causality, and the mode of representation of internal processes;
specific examples of each of these different response types are
presented in Online Resource 1. The 200 matched control stories are not
represented in this particular table of feature-wise agreements.

\begin{longtable}[]{@{}
  >{\raggedright\arraybackslash}p{(\columnwidth - 6\tabcolsep) * \real{0.4}}
  >{\raggedright\arraybackslash}p{(\columnwidth - 6\tabcolsep) * \real{0.24}}
  >{\raggedright\arraybackslash}p{(\columnwidth - 6\tabcolsep) * \real{0.17}}
  >{\raggedright\arraybackslash}p{(\columnwidth - 6\tabcolsep) * \real{0.19}}@{}}
\caption{Exact agreement between StoryScope and author-reviewed feature values in 200 focal stories}\label{tab:7}\\
\toprule\noalign{}
\begin{minipage}[b]{\linewidth}\raggedright
\textbf{Feature (code)}
\end{minipage} & \begin{minipage}[b]{\linewidth}\raggedright
\textbf{Narrative dimension}
\end{minipage} & \begin{minipage}[b]{\linewidth}\raggedright
\textbf{Comparable stories}
\end{minipage} & \begin{minipage}[b]{\linewidth}\raggedright
\textbf{Exact agreement}
\end{minipage} \\
\midrule\noalign{}
\endfirsthead
\toprule\noalign{}
\begin{minipage}[b]{\linewidth}\raggedright
\textbf{Feature (code)}
\end{minipage} & \begin{minipage}[b]{\linewidth}\raggedright
\textbf{Narrative dimension}
\end{minipage} & \begin{minipage}[b]{\linewidth}\raggedright
\textbf{Comparable stories}
\end{minipage} & \begin{minipage}[b]{\linewidth}\raggedright
\textbf{Exact agreement}
\end{minipage} \\
\midrule\noalign{}
\endhead
\bottomrule\noalign{}
\endlastfoot
Dominant narrative person (F01) & Perspective & 200 & 85.5\% \\
Breadth of focalization across characters (F02) & Perspective & 200 &
50.5\% \\
Institutions as Social Actors (F03) & Social Networks & 200 & 59.0\% \\
Range of Relationship Types Present (F04) & Social Networks & 200 &
2.5\% \\
Primary technique for rendering inner life (F05) & Agents & 200 &
35.0\% \\
Protagonist initiative vs. reactivity (F06) & Agents & 200 & 28.5\% \\
Explicitness of causal links (F07) & Events & 200 & 29.0\% \\
Clear Single Climactic Event (F08) & Events & 200 & 79.5\% \\
Global Arc Shape (F09) & Plot & 200 & 77.0\% \\
Degree of Closure (F10) & Plot & 200 & 34.5\% \\
Global Withholding Intensity (F11) & Revelation & 199 & 27.6\% \\
Resolution Explicitness of Core Questions (F12) & Revelation & 199 &
54.8\% \\
Global Chronological Structure (F13) & Temporal Structure & 200 &
60.5\% \\
Degree of Chronological Discontinuity (F14) & Temporal Structure & 200 &
62.0\% \\
\end{longtable}

Note: The denominator is 200 focal stories except for the two Revelation
features (199 with both values available). Exact agreement requires
identical values or, for the multi-select feature, an identical full set
of selected answers. This is a retrospective audit of preserved
author-reviewed values, not an independent annotation exercise.
Additional response-type and dimension summaries and frequent
disagreement categories are in Online Resource 1.

\section{5. Model-pair robustness and
heterogeneity}\label{model-pair-robustness-and-heterogeneity}

The mean masked RCI across the seven narrative dimensions is positive
for all ten model pairs examined, ranging from .372 to .440 (Figure 2).
The highest values are observed for Kimi-Gemini (.440), DeepSeek-Kimi
(.438), and GPT-Claude (.426), while the lowest is observed for
Claude-Gemini (.372). These descriptive differences do not, by
themselves, establish their underlying causes.

\begin{figure}[htbp]
\centering
\includegraphics[width=.90\linewidth]{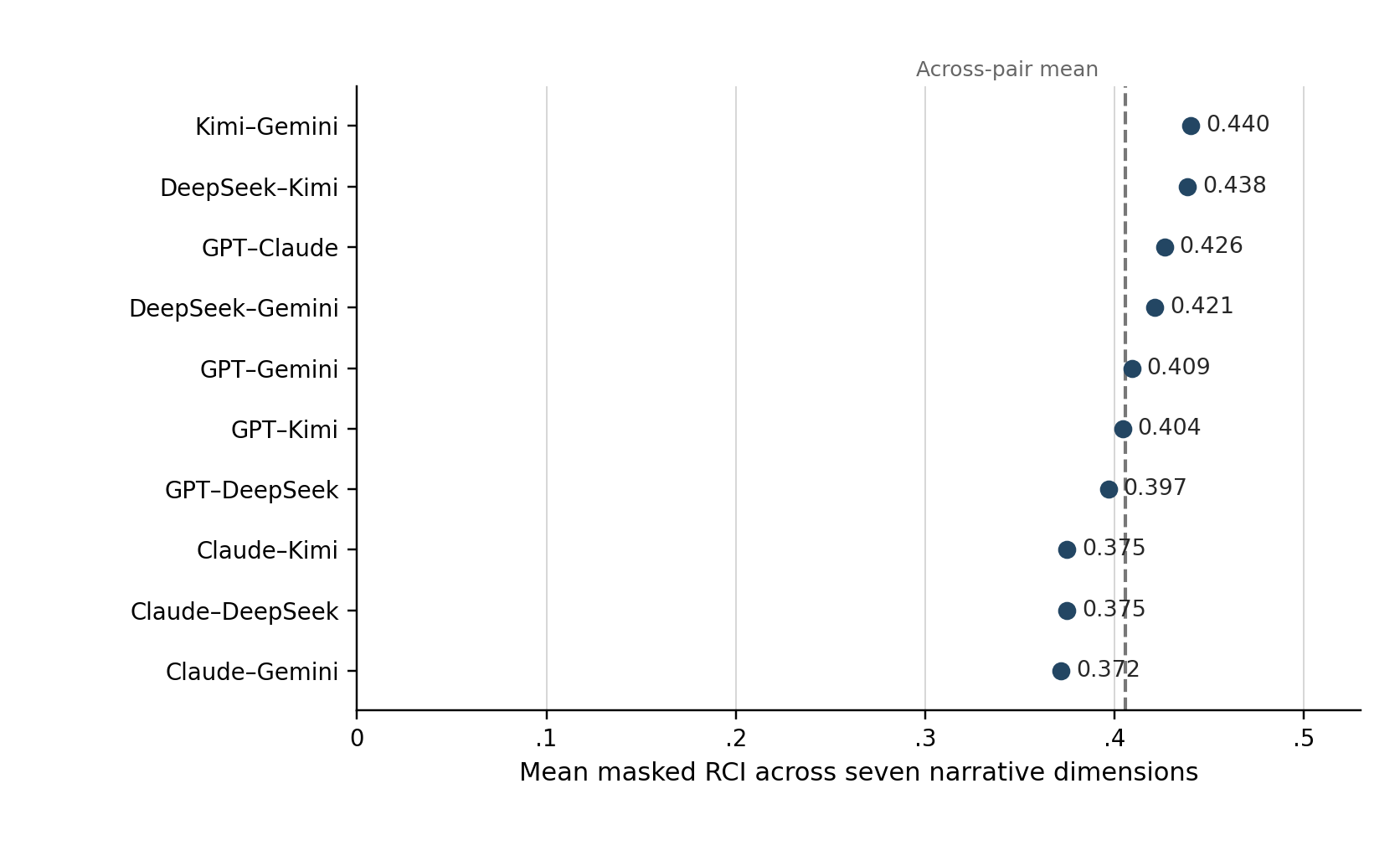}
\caption{Mean masked RCI across seven narrative dimensions for each model pair. The dashed line marks the mean across pairs. Positive RCI estimates occur across all model pairs rather than being concentrated in one relationship. The horizontal axis begins at zero, and each point is labeled with its RCI value to aid descriptive comparison.}\label{fig:2}
\end{figure}

Furthermore, the masked RCI is positive for all 70
model-pair-by-dimension combinations, ranging from .299 (Claude-Gemini,
Temporal Structure) to .547 (Kimi-Gemini, Perspective). These estimates
are not independent because they share the same five model generators,
prompts, and StoryScope feature extraction. No correction for multiple
comparisons is applied, as these estimates are reported descriptively
rather than as results of separate significance tests.

\section{6. Discussion: implications for evaluation of conditional
generation}\label{discussion-implications-for-evaluation-of-conditional-generation}

Overall, the matched-control, masking, and recoding analyses reveal a
positive same-prompt versus matched-prompt distance contrast whose
magnitude depends on the controls used and the way narrative features
are measured. These results apply only to the five specific models
examined, not to all language models. The semantic mask is less
consistent with the author\textquotesingle s reference labels near its
decision boundary: 53 of the 76 disagreements occur in this region,
where accuracy is .621.

Even though mean RCI varies very little between .409 and .406 due to
masking, this does not imply that the prompt influence is small since
the mask eliminates those features of the prompt which are considered
most semantically related to the prompts. The estimated value of .250
with the help of .75 caliper includes just one matching neighbor and 690
selected prompts, while the main value of .406 involves three neighbors
and 10,248 prompts. It is impossible to assume that their difference
lies in the similarity threshold. This further emphasizes the need for
baselines in discussions about output homogeneity (Jiang et al., 2025;
Schaeffer et al., 2026, preprint). Positive RCI is the comparison of
same-prompt versus matched-prompt distances, not the model behavior
without prompts.

While the author audit finds a negative cosine-adjusted distance
coefficient in its small sample, the independently selected expanded
audit is unable to replicate an association of similar magnitude or
statistical significance. Given that alignment scores are associated
with embedding similarity and that there are only a few focal prompts
with controls in both alignment groups, these findings shed light on the
shortcomings of the observational baseline.

Aside from prompt-based controls, the technique used for calculating the
feature measures can influence the size of the estimate. In particular,
the full-corpus .406 and the recoding .187 have differences in sample
size, in which features are included, and in aggregation. Thus, they
should not be seen as an example of before-and-after comparison.

The direct measure experiment controls for the number of 40 prompt sets,
controls based on nearest-neighborhood, comparison model, 14 feature
types, distance computation rules, and aggregation technique (mean
prompt-level RCI): .380 with StoryScope values and .187 with Author
Reviewed Values of Annotator A (Section 4.7).This difference doesn't
prove that one set of labels describes the stories better than the
other; additional research is required to analyze why feature value
differences lead to distance differences. The sensitivity to measurement
emphasizes the importance of distinguishing a construct from its
numerical representation (Jacobs \& Wallach, 2021; Moretti, 2013) and
validating text-analysis tools for particular tasks (Grimmer \& Stewart,
2013).

Feature-based analysis helps identify potential sources of measurement
sensitivity. In regard to the focal story sample, high congruence can be
observed regarding dominant narrative person and the presence of a
single climax, while the level of congruence is lower in relation to
protagonist agency, the explicitness of causality, and withholding
intensity. For instance, the code "mostly initiates key actions," as
defined by the author, often corresponds to the code "mixed initiative
and reactivity," as assessed in StoryScope, while the code
"predominantly implicit" causality, as defined by the author, often
corresponds to the code "mixed explicit and implicit" in StoryScope.
These discrepancies may impact the distances of both same-prompt and
matched-control samples, but the present agreement audit does not
establish which feature discrepancies contribute quantitatively to the
overall reduction in RCI values.

\section{7. Limitations and threats to
validity}\label{limitations-and-threats-to-validity}

This study has several key limitations. First, it examines only five
specific AI models and does not include independently written human
stories. To compare stories generated by AI models with those written by
humans under the same conditions, it would be necessary to collect
human-written stories produced in response to the same prompts. Such
stories were not available in the dataset. Second, the semantic mask
shows substantial disagreement with the reference labels near its cutoff
(accuracy = .621), and the stratified audit cannot determine the
prevalence of label categories in the corpus.

Third, the annotation procedures involved the author as Annotator A and
included preliminary AI-assisted labeling. Although Online Resource 1
preserves the reference labels used for analysis, the original coding
sheets and the complete record of the author\textquotesingle s
annotation and review process have not been preserved. This hinders
independent evaluation of the original labeling procedure; the reported
validation and recoding analyses use the retained reference values.
Fourth, the analysis uses the definitions and extracted feature values
from StoryScope. In the direct comparison using the same 40 focal
prompts, matched controls, and 14 features, the author-reviewed coding
produces a lower but still positive RCI than the original StoryScope
values, although both use the same taxonomy. The recoded dimension-level
results also show less consistent positive estimates. However, the
recoding does not allow us to test alternative narrative classification
schemes or rank the narrative dimensions.

Finally, the matched-prompt baseline is observational: semantically
similar prompts are not experimentally equivalent. By narrowing the
caliper for similarity, different prompts are included in the sample;
the subset at the .75 threshold is just 6.7\% of the total corpus, and
any attempt to extrapolate beyond .80 is unwarranted. In both constraint
audits, the ratings are based on author evaluations and selected
samples. There are few focal prompts with controls from both alignment
groups in both audits, and there is an association between alignment and
embedding similarity. Conditional bootstrap intervals do not incorporate
uncertainty in the ratings or the original neighbor-selection procedure.
These limitations prevent a generalizable estimate of an independent
narrative-constraint alignment effect.

\section{8. Conclusion}\label{conclusion}

A total of 51,240 StoryScope stories were examined based on same-prompt
and semantically matched-prompt distances, 420 prompt-feature pairs were
checked, and 400 story records were recoded. The results of the first
main analysis showed positive RCI scores for all seven dimensions using
the 213-feature data set. In terms of the single-neighbor robustness
analysis, more stringent similarity criteria were linked to lower RCI
scores for successively reduced prompt data sets. For the direct
measurement comparison with the same 40 focal prompts, 14 features, and
identical aggregation approach, the average prompt-level RCI was .380
using StoryScope values and .187 using author-reviewed values,
respectively.

The contribution made by this research is the demonstration of a
procedure for assessing replicable conditional narrative similarity.
This procedure involves the use of semantically matched prompts, an
evaluated prompt-relevance mask, and comparing alternative feature
assignments. The findings from the five specific models show that there
is a positive difference between same-prompt and matched-prompt
distances. Nonetheless, this difference is contingent on the method of
matching prompts and measuring the narrative components. Thus, a
positive RCI can be considered a signal of convergence relative to the
chosen controls, but not a prompt-independent trait of model-generated
narratives. Exploratory audits show that prompts close in embedding
space can impose different narrative requirements. However, the expanded
cosine-controlled audit does not establish a separate
constraint-alignment effect.

\section{Statements and Declarations}\label{statements-and-declarations}

\textbf{Competing interests.} The author has no financial or
non-financial competing interests to declare.

\textbf{Ethics approval.} This research analyzed publicly available
textual data. The author served as the sole human annotator, evaluating
textual features rather than participating as a research subject. No
human participants were recruited as research subjects, and no
experiments involving animals were conducted.

\textbf{Generative AI use.} This study used generative AI tools to
assist with writing and debugging the analysis code, linguistic editing
of the manuscript, and preparing preliminary labels for prompt--feature
validation and story--feature recoding. The author was the sole
annotator as described in the Methods section.

\textbf{Code and materials availability.} The code needed to reproduce
the study\textquotesingle s analysis, including the analysis manifests,
validation files, results, and checksums, is provided in Online Resource
1. The original StoryScope code is freely available in the
project\textquotesingle s GitHub repository. The provided ZIP archive
has the following SHA-256 checksum:

6bc0d8611b990970f3a770708fbfd9badb72c408b0ed2c09ba789426705a3f45

No version-control commit ID is provided for the code developed
specifically for this study.

\section{Supplementary Information}\label{supplementary-information}

Online Resource 1 (ESM\_1.zip) includes the corpus file manifests,
feature lists, saved primary distances, missing-value examinations,
semantic-mask validation results, Annotator A\textquotesingle s retained
reference labels and their provenance, recoding results, and comparisons
of alternative feature measurements using matched prompts.

This resource also comprises the initial analysis notebook, distances at
the neighborhood level, bootstrap results, and the investigation of
dependency due to common prompt IDs. The additional README file
separates the initial 600 samples of automatic masks from the final 420
samples of author annotations.

\section{References}\label{references}

Anderson, B. R., Shah, J. H., \& Kreminski, M. (2024). Homogenization
effects of large language models on human creative ideation. In
Proceedings of the 16th Conference on Creativity and Cognition (pp.
413-425). Association for Computing Machinery.
\url{https://doi.org/10.1145/3635636.3656204}

Chatman, S. (1978). Story and discourse: Narrative structure in fiction
and film. Cornell University Press.
\url{https://utpdistribution.com/9780801411311/story-and-discourse/}

Doshi, A. R., \& Hauser, O. P. (2024). Generative AI enhances individual
creativity but reduces the collective diversity of novel content.
Science Advances, 10(28), eadn5290.
\url{https://doi.org/10.1126/sciadv.adn5290}

Genette, G. (1980). Narrative discourse: An essay in method (J. E.
Lewin, Trans.). Cornell University Press.
\url{https://search.worldcat.org/title/Narrative-discourse-\%3A-an-essay-in-method/oclc/1468005412}

Gilardi, F., Alizadeh, M., \& Kubli, M. (2023). ChatGPT outperforms
crowd workers for text-annotation tasks. Proceedings of the National
Academy of Sciences, 120(30), e2305016120.
\url{https://doi.org/10.1073/pnas.2305016120}

Gower, J. C. (1971). A general coefficient of similarity and some of its
properties. Biometrics, 27(4), 857--871.
\url{https://doi.org/10.2307/2528823}

Grimmer, J., \& Stewart, B. M. (2013). Text as data: The promise and
pitfalls of automatic content analysis methods for political texts.
Political Analysis, 21(3), 267--297.
\url{https://doi.org/10.1093/pan/mps028}

Jacobs, A. Z., \& Wallach, H. (2021). Measurement and fairness. In
Proceedings of the 2021 ACM Conference on Fairness, Accountability, and
Transparency (pp. 375--385). Association for Computing Machinery.
\url{https://doi.org/10.1145/3442188.3445901}

Jiang, L., Chai, Y., Li, M., Liu, M., Fok, R., Dziri, N., Tsvetkov, Y.,
Sap, M., \& Choi, Y. (2025). Artificial hivemind: The open-ended
homogeneity of language models (and beyond). Advances in Neural
Information Processing Systems, 38 (Datasets and Benchmarks Track).
\url{https://proceedings.neurips.cc/paper_files/paper/2025/hash/754d5a526a5ee5a47220664a0eb92751-Abstract-Datasets_and_Benchmarks_Track.html}

Ma, Y., Shi, K., He, H., Zhang, Z., Lei, Z., Qiu, Z., Hu, R., \& Liu, J.
(2026). Creative convergence or imitation? Genre-specific homogeneity in
LLM-generated Chinese literature. arXiv.
\url{https://doi.org/10.48550/arXiv.2603.14430}

Moretti, F. (2013). ``Operationalizing'': Or, the function of
measurement in literary theory. New Left Review, 84, 103--119.
\url{https://newleftreview.org/issues/ii84/articles/franco-moretti-operationalizing}

Padmakumar, V., \& He, H. (2024). Does writing with language models
reduce content diversity? International Conference on Learning
Representations.
\url{https://proceedings.iclr.cc/paper_files/paper/2024/hash/02dec8877fb7c6aa9a79f81661baca7c-Abstract-Conference.html}

Piper, A. (2018). Enumerations: Data and literary study. University of
Chicago Press.
\url{https://doi.org/10.7208/chicago/9780226568898.001.0001}

Plank, B. (2022). The ``problem'' of human label variation: On ground
truth in data, modeling and evaluation. In Proceedings of the 2022
Conference on Empirical Methods in Natural Language Processing (pp.
10671--10682). Association for Computational Linguistics.
\url{https://doi.org/10.18653/v1/2022.emnlp-main.731}

Rathje, W. (2026). Computational narratology with transformer
embeddings. Computational Humanities Research, 2, e15.
\url{https://doi.org/10.1017/chr.2026.10035}

Reimers, N., \& Gurevych, I. (2019). Sentence-BERT: Sentence embeddings
using Siamese BERT-networks. In Proceedings of the 2019 Conference on
Empirical Methods in Natural Language Processing and the 9th
International Joint Conference on Natural Language Processing (pp.
3982--3992). Association for Computational Linguistics.
\url{https://doi.org/10.18653/v1/D19-1410}

Rettberg, J. W., \& Wigers, H. (2025). AI-generated stories favour
stability over change: Homogeneity and cultural stereotyping in
narratives generated by GPT-4o-mini. Open Research Europe, 5, 202.
\url{https://doi.org/10.12688/openreseurope.20576.1}

Roberts, M. E., Stewart, B. M., \& Nielsen, R. A. (2020). Adjusting for
confounding with text matching. American Journal of Political Science,
64(4), 887--903. \url{https://doi.org/10.1111/ajps.12526}

Russell, J., Rajendhran, R., Iyyer, M., \& Wieting, J. F. (2026).
StoryScope: Investigating idiosyncrasies in AI fiction. In Proceedings
of the Third Conference on Language Modeling (COLM 2026,
pre-proceedings).
\url{https://proceedings.colm.cc/paper_files/paper/2026/hash/afb715947210f5656cbe8917fb8a55ce-Abstract-Conference.html}

Schaeffer, R., Miranda, B., Kazdan, J., Chudnovsky, J., \& Koyejo, S.
(2026). How strong is the evidence for the artificial hivemind?
Reevaluating evidence for the open-ended homogeneity of language models
{[}Preprint{]}. arXiv. \url{https://arxiv.org/abs/2609.33936}

Stuart, E. A. (2010). Matching methods for causal inference: A review
and a look forward. Statistical Science, 25(1), 1--21.
\url{https://doi.org/10.1214/09-STS313}

Tevet, G., \& Berant, J. (2021). Evaluating the evaluation of diversity
in natural language generation. In Proceedings of the 16th Conference of
the European Chapter of the Association for Computational Linguistics:
Main Volume (pp. 326--346). Association for Computational Linguistics.
\url{https://doi.org/10.18653/v1/2021.eacl-main.25}

Tian, Y., Huang, T., Liu, M., Jiang, D., Spangher, A., Chen, M., May,
J., \& Peng, N. (2024). Are large language models capable of generating
human-level narratives? In Proceedings of the 2024 Conference on
Empirical Methods in Natural Language Processing (pp. 17659-17681).
Association for Computational Linguistics.
\url{https://doi.org/10.18653/v1/2024.emnlp-main.978}

van der Lee, C., Gatt, A., van Miltenburg, E., Wubben, S., \& Krahmer,
E. (2019). Best practices for the human evaluation of automatically
generated text. In Proceedings of the 12th International Conference on
Natural Language Generation (pp. 355--368). Association for
Computational Linguistics. \url{https://doi.org/10.18653/v1/W19-8643}

Wenger, E., \& Kenett, Y. N. (2026). Large language models are
homogeneously creative. PNAS Nexus, 5(3), pgag042.
\url{https://doi.org/10.1093/pnasnexus/pgag042}

Xu, W., Jojic, N., Rao, S., Brockett, C., \& Dolan, B. (2025). Echoes in
AI: Quantifying lack of plot diversity in LLM outputs. Proceedings of
the National Academy of Sciences, 122(35), e2504966122.
\url{https://doi.org/10.1073/pnas.2504966122}

\end{document}